\documentclass{article}

\usepackage[final, nonatbib]{neurips_2023}
\usepackage[numbers]{natbib}
\usepackage{algorithm}
\usepackage{algpseudocode}
\usepackage{xspace}
\usepackage{subcaption}
\usepackage{pgf}
\usepackage{tabularx}
\usepackage{booktabs}
\usepackage{amssymb}
\usepackage{amsmath}

\newcommand{\said}{\ensuremath{\mathcal{S}_t}\xspace}
\newcommand{\heard}{\ensuremath{\mathcal{H}_t}\xspace}

\title{Learning interactions to boost human creativity with bandits and GPT-4}

\author{
  Ara Vartanian \\
  University of Wisconsin-Madison \\
  \texttt{aravart@cs.wisc.edu} \\
  \And
  Xiaoxi Sun \\
  University of Wisconsin-Madison \\
  \texttt{xsun279@wisc.edu} \\
  \And
  Yun-Shiuan Chuang \\
  University of Wisconsin-Madison \\
  \texttt{yunshiuan.chuang@wisc.edu} \\
  \And
  Siddharth Suresh \\
  University of Wisconsin-Madison \\
  \texttt{siddharth.suresh@wisc.edu} \\
  \And
  Xiaojin Zhu \\
  University of Wisconsin-Madison \\
  \texttt{jerryzhu@cs.wisc.edu} \\
  \And
  Timothy T. Rogers \\
  University of Wisconsin-Madison \\
  \texttt{ttrogers@wisc.edu} \\
}

\begin{document}

\maketitle

\begin{abstract}
  This paper considers how interactions with AI algorithms can boost human creative 
  thought. We employ a psychological task that demonstrates limits on
  human creativity, namely semantic feature generation: given a concept name,
  respondents must list as many of its features as possible. Human participants
  typically produce only a fraction of the features they know before getting
  ``stuck.'' In experiments with humans and with a language AI (GPT-4) we contrast
  behavior in the standard task versus a variant in which participants can ask
  for algorithmically-generated hints. Algorithm choice is administered by a
  multi-armed bandit whose reward indicates whether the hint helped generating
  more features. Humans and the AI show similar benefits from hints, and
  remarkably, bandits learning from AI responses prefer the same prompting
  strategy as those learning from human behavior. The results suggest that
  strategies for boosting human creativity via computer interactions can be
  learned by bandits run on groups of simulated participants.
\end{abstract}

\section{Introduction}

In an early study of human problem solving, Maier asked participants to tie
together two ropes hanging from the ceiling~\cite{maier1931reasoning}. The ropes were far enough apart that it was impossible to reach one while holding on to the other, but participants made creative use of various objects strewn about the room: lengthening one rope by affixing an extension cord, for instance, or hooking the second rope on a broom handle. Each time a solution was reached, the experimenter asked for a new one. Eventually the obvious solutions were depleted, at which point the experimenter, seemingly bored, would bat lightly at one of the ropes. With remarkable frequency, participants would quickly hit upon a new solution: tie an object to one rope, swing it like a pendulum, run to grasp the second, and catch the first as it swung back. Participants typically had little idea how they hit upon the answer, and made no reference to Maier's subtle hint when asked.

This classic study highlights two interesting properties of human thought: first, though human minds are creative, they can also easily get stuck: as every scientist has experienced, it can be remarkably difficult to generate new ideas that then seem obvious in retrospect. Second, these mental roadblocks can sometimes be overcome by environmental cues, even when the cues themselves seem relatively opaque. Indeed, despite almost a century and dozens of studies since Maier's paper, it remains unclear what kinds of environmental cues will help people overcome mental roadblocks when generating new ideas~\cite{robertson2016problem}.

The current work assesses whether human interactions with adaptive AI algorithms can serve as a kind of ``creative prosthesis'' in such situations. Consider a scenario in which, after much thought, a human agent has run out of ideas and asks an AI agent for help.
 The AI agent has access to all ideas produced thus far, and possess many different computer algorithms for automatically generating hints that could potentially help the user think of new idea. It is not clear {\it a-priori}, however, which hint-generating algorithms are likely to be effective. How then might the AI agent decide which hint-generating algorithm to use at a given moment?

Our main idea is to treat this task as a multi-armed bandit (MAB) problem. The AI agent is a $k$-armed bandit.
Each arm is a hint-generating algorithm. Upon pulling an arm -- i.e. deploying a given algorithm to produce a hint -- the user may or may not be able to generate new ideas.
The quantity (and quality) of new ideas is the immediate reward given to that arm.
The bandit will then follow its exploration vs. exploitation strategy in order to maximize expected cumulative rewards, which is to help the user generate many new ideas.
Study 1 evaluates this approach using experiments in which the users are human participants.
The central question is whether such interactions leads the bandit algorithm to favor arms
that reliably improve human performance in the cognitive task. 

Training a bandit on human behavior is, however, prohibitively expensive in many applications where the
number of potential hint algorithms increases to hundreds or thousands.
Study 2 evaluates whether important patterns of human behavior from Study 1 are captured
by simulations with a contemporary large language model (specifically GPT-4)
~\cite{openai2023gpt4}, and if so, whether the bandit can be pre-trained using
GPT-4 behavior as a proxy for human interaction. If bandits trained in this way learn
to prefer the same hinting strategies as those trained directly on human behavior, this
suggests that language models can be used to ``warm the seat'' to find promising
hinting strategies for human users amongst a much broader pool of possibilities.

Toward these ends, we employed a simple generation task commonly studied in cognitive science and known to reliably produce mental roadblocks, namely the {\it verbal fluency} task in which participants must produce as many appropriate responses to a cue as they can think of~\cite{shao2014verbal,mathuranath2003effects}. For instance, they may be asked to list all the animals they know, or all the words they can think of beginning with F. For reasons that remain poorly understood, people produce only a small fraction of the responses they know in such tasks before running out of ideas: despite knowing hundreds of animal names, most people can produce just 20-30 before running dry~\cite{jun2015human,hills2012optimal}. Our studies used a {\it feature fluency} task ~\cite{mcrae2005semantic}, where participants must list all the properties of a familiar concept such as {\it penguin} or {\it journalist.} As with other fluency tasks, people know many more properties than they can produce before getting stuck, but feature fluency additionally affords a greater degree of creativity/flexibility since participants are free to generate a wide variety of different response types. 

Study 1 evaluates whether a bandit AI agent can reliably improve human performance on the feature-generation task.
The bandit interacts individually with each human participant, learning strategy preferences
independently for each. We consider (a) whether hints provided by the bandit AI lead people to generate more features, and more diverse features, relative to unhinted conditions, (b) whether and how features generated relate semantically to hints provided, and (c) whether the bandit learns to prefer some hinting strategies over others.

Study 2 replicates Study 1, but uses GPT-4 to simulate human performance in the feature-generation task.
Note that GPT-4 does not possess the same cognitive constraints on memory search as
humans, and from extensive training on vast amounts of language, has the potential to exhibit
super-human performance on the feature-listing task. On the other hand, several studies have
found that the model can generate remarkably human-like patterns of responding in some
tasks~\cite{dillion2023can}, while others have demonstrated a tendency to ``hallucinate'' facts that
are incorrect. Thus a key and nontrivial question is whether GPT-4 captures 
important aspects of human behavior in the task when given analogous instructions in both
hinted and unhinted conditions. Following this analysis, we consider whether bandits
interacting with GPT-4 can learn strategies that reliably boost its performance; whether
GPT-4 responses relate semantically to hints provided to the bandit; and how strategy preferences
learned by the bandit relate to those learned from human behavior. In the discussion we
consider the implications of this work for augmenting human thought at much larger scale, and 
for more difficult problems. 

\section{Related Work}

Verbal fluency has been extensively studied in cognitive science, both behaviorally~\cite{rosen1980verbal,regard1982children,whiteside2016verbal}
and computationally~\cite{zemla2017modeling,jun2015human,hills2012optimal}. A rich psychology literature has also characterized a range
of problem-solving phenomena where people struggle to find creative solutions ~\cite{robertson2016problem}
and outlines several hypotheses about how they eventually achieve insight ~\cite{ollinger2009psychological}.
Likewise, human-AI co-creation has been explored in a number 
of domains including drawing~\cite{karimi2020creative}~\cite{zhang2022storydrawer}, storytelling
~\cite{yang2022ai}, and design~\cite{kim2021studying}. To our knowledge, however,
this is the first work exploring Human-AI interaction specifically in
facilitating performance in the verbal fluency task, or indeed any task where humans are prone to
mental roadblocks.~\cite{lehman2011abandoning} and
~\cite{liapis2023designing} have argued for playfulness- or novelty-based
objectives in co-creativity tasks. Much of the vast literature on prompting
methods for large language~\cite{liu2023pre} is not immediately applicable to
this work due to the open-ended nature of the feature recall problem and
adaptation to the session history.

\section{Study 1: Multi-Armed Bandit for Human Creativity}

\subsection{Computational Method: The Bandit}

Our AI agent is the standard $k$-armed adversarial bandit algorithm EXP3~\cite{auer2002nonstochastic}, which balances exploration and exploitation, makes only weak assumptions (in particular, no stochastic assumption on arm rewards), and enjoys a sublinear regret guarantee.
We model different hint generating algorithms as bandit arms, and will specify them below.
Whenever a user wants a hint, it is an arm pull request to the bandit.
The bandit chooses which arm to pull based on the entire history of user interaction. 
If the resulting hint causes the user to generate new features, the arm is rewarded.
To follow adversarial bandit convention, we use loss (negation of reward) to express the EXP3 algorithm.
Specifically, we use binary arm loss $\ell_{t,a_t}=0$ if the user produces one or more new features after the $t$-th hint produced by the $a_t$-th hint algorithm; otherwise $\ell_{t,a_t}=1$.
We present the EXP3 algorithm with interpretation for our human experiment in \textbf{Algorithm~\ref{alg:exp3}}.
\begin{algorithm}
  \caption{EXP3 Algorithm Adapted to Feature Recall Protocol}
  \label{alg:exp3}
  \begin{algorithmic}[1]
    \State Initialize weights  $w_{11} = \ldots = w_{1k} = 1$
    \For{$t = 1, 2, \ldots $}
    \State Wait until the user requests a hint
    \State Compute arm probability $p_{t,i} = {w_{t,i} \over \sum_{j=1}^k w_{t,j}}$ for all $i \in [k]$
    \State Sample arm $a_t \sim p_t$
    \State Use the $a_t$-th hint-generating algorithm to generate a hint, give the hint to the user, receive arm loss $\ell_{t,a_t}$.
    \State Update arm weights with inverse probability weighting $w_{t+1,i} = w_{t,i} \exp\left(-\eta {1_{[a_t=i]} \over p_{t,i}}\ell_{t,i}\right)$ for all $i \in [k]$.
    \EndFor
  \end{algorithmic}
\end{algorithm}

Here $t$ indexes the number of hints, not the number of features the participant has generated. 
For example, at $t=1$ EXP3 generates a hint to the user; the user may subsequently produce ten features from that hint; but the next time the user requests a hint, $t$ will be 2 instead of 11.
This interaction proceeds for the duration of the participant's session.
The indicator function $1_{[a_t=i]}$ is 1 if $a_t=i$ and 0 otherwise.
For standard no-regret analysis, the step size parameter $\eta$ is optimized as $\eta =
\sqrt{\frac{2 \ln k}{T k}}$, where $T$ is the assumed horizon (total number of hints).
Before our human experiments, we estimate a horizon $T = 20$, giving $\eta \approx 0.19$.
We use this $\eta$ throughout the experiments.

\subsection{Computational Method: The Arms}

We experimented with $k=3$ arms.
Each arm, when pulled, produces a hint consisting of a list of 5 English words.
The arms differ in the algorithm that chooses those 5 words based on user interaction history and external resources such as word embedding.
Importantly, \emph{a priori} we do not claim any arm is useful for improving human creativity.
This is the benefit of our bandit framework: the true value of the arms are learned via interactions.
Our framework also allows one to easily add / replace more arms.

In the following arm descriptions, embeddings $\phi(\cdot)$ were computed using
\texttt{fastText}~\cite{bojanowski2016enriching} while distributional data was
provided by~\cite{segaran2009beautiful}. The candidate vocabulary $V$ refers to
the intersection of vocabularies defined in \texttt{fastText} and
~\cite{segaran2009beautiful}. At iteration $t$, \heard is the set of word types
occurring in provided hints throughout history, while \said is the set of word types occurring in
the participant's produced features throughout history. Word types $w$ can be
associated with the frequency statistic $f_w$ provided by~\cite{segaran2009beautiful}.

\textbf{Arm 1: Semantic Neighbors.}
When pulled, arm 1 looks at what feature words the user has produced so far and basically gives 5 neighboring words in the word embedding space. 
For example,
suppose the user only produced the feature word ``penguin'' up to this point, then arm 1 may produce the hint ``dolphin mammals species ships whaling.'' 
The rationale is that semantically related words may help users further expand their feature listing.
However, when the user has already produced many features, it will be important to judiciously choose one of those historical words to generate the neighbors.
Arm 1 chooses the historical word with the least frequency in general English.
This tends to choose the most salient historical word,
and leads the user to parts of the semantic space that are less easily accessible~\cite{jescheniak1994word}.
Formally,
at arm pull $t$, arm 1 constructs
a hint by choosing the word type $c_t \in \said$ which is the least frequently according to $f_w$,
and has not yet been chosen in any previous arm 1 pulls.
 The 5 nearest neighbors of $\phi(c_t)$ among $\phi(V)$ are concatenated to
form the hint.

\textbf{Arm 2: English Frequency.}
This arm simply samples 5 words 
from the English vocabulary
based on large corpus word frequency $f_w$. On each hint round, words previously produced by
the participant or used in prior hints are removed from the vocabulary and five
hints are drawn without replacement from the frequency-based multinomial
distribution over remaining words. More formally, we sample $w \sim f_w$ for $v \in V
\cap \heard^C$.

Because they are randomly drawn, hint words bear no principled relationship to the target concept, prior responses, or prior hint words -- for instance, one hint for the concept {\it journalist} was ``search rocket centre incorporating point.'' The rationale for this hint strategy arises from the hypothesis that people become mentally ``stuck'' in a semantic neighborhood, unable to think of semantically relevant features that they have not already produced~\cite{troyer1997clustering}. If this is so, provision of random cues may push people out of the saturated part of the space, allowing them to find new, relevant features.

\textbf{Arm 3: Diversity Cover.}
This arm also aims to ``unstick'' the user, but in a different way than Arm 2.
It tries to cover the whole English semantic space quickly, by 
 sampling hint words that are semantically distal to all previously user-produced words and hints. 
Consider all the word types appearing in prior feature responses \said and previously-generated hints $\heard$.
The remaining English word types form a point cloud in the word embedding space.
Arm 3 tries to maximally spreads out the 5 hint words, with a stochastic procedure, to cover this point cloud.
Formally, the point cloud to be covered is $V \setminus (\said \cup \heard)$.
Each word type $v_i \in V \setminus (\said \cup \heard)$ is assigned a probability $p_i$ proportional to its minimum squared distance to the historical words: $p_i \propto \min_{w \in {\said \cup \heard}} d(\phi(w), \phi(v_i))^2$.
The choice of squared distance is motivated by the kmeans++ algorithm~\cite{arthur2007k}.
The intuition is that $v_i$ has a larger probability of being sampled if it is far from all historical words.
Arm 3 then samples 5 hint words from $\{p_i\}$.

Because hint words are semantically distal to one another and to the produced
responses, increasing numbers of hints are guaranteed to maximally explore
different regions of the semantic space. As an example, one hint produced by
this strategy for the concept {\it journalist} was ``cybersecurity vary
anchoring twitter reservoirs''. The rationale for this strategy is that it
ensures hint words express a very diverse set of possible meanings, rendering it
more likely that new hints will eventually find regions of the space that help
the participant generate new responses.

\subsection{Human Behavioral Experiments}

{\em Participants.} 37 undergraduates at the University of Wisconsin-Madison completed the study for partial course credit. The study was approved by the UW IRB board for the Social and Behavioral Sciences.

{\em Procedure.} 
Participants completed the study online through a web interface that presented them with the name of a concept and asked them to generate as many of its properties as they could (see \textbf{Figure~\ref{fig:instructions}}). 
In the {\it hinted} condition, instructions indicated that participants could request a hint by pressing a button in the user interface.
This initiates an arm pull in the bandit algorithm.
Hint words returned by the bandit algorithm would then be displayed in the interface. Participants were further instructed that hints may or may not be useful, and that they could request a new hint at any time. In the unhinted condition, the interface text indicated that no hints were available.  Participants entered each feature property by typing a short phrase into a text box and hitting Enter; for instance, properties of the concept {\it penguin} might include phrases like ``is black and white,'' ``has feathers,'' or ``appeared in the movie Happy Feet.'' Each such phrase was counted as one response. The study began with a short practice session using two unrelated concepts ({\it tiger} and {\it desk}) to familiarize respondents with the interface. 
In the experiment proper, all participants completed two 20 minute sessions, one using the concept {\it penguin} and the other using the concept {\it journalist}. For each participant, one concept appeared in the {\it hinted} condition and the other the {\it unhinted} condition. Pairing of concept and condition, and the ordering of the two tasks, was counterbalanced across participants.

\subsection{Results}

One participant produced a very large number of features, more than 3.5 standard deviations above the sample mean, including a large number of highly arbitrary features (for instance, listing drinking every possible variety of tea as properties of {\em journalist}). This participant's data was removed and analyses were conducted on the remaining 36.

\textbf{Do hints lead participants to produce more features?} 
Participants produced a median of 34 features in the
hinted condition and 26.5 in the unhinted condition. 
To evaluate whether these differences are
statistically reliable, we fit a series of nested mixed effects
models predicting the number of features produced by each participants
from the condition (hinted/unhinted), the concept (penguin/journalist), 
and the block order (first block/second block). 
\textbf{Table~\ref{tab:model_comp}} shows fit statistics and comparisons of
all models in the series. The best-fitting model showed a significant fixed effect of 
condition, with reliably more features produced in hinted than unhinted condition
(model $m: \beta=8.0, p < 0.001$ vs null model $m_0$), and a significant fixed effect of block order, 
with reliably fewer features produced in the second block ($m3: \beta=-5.1, p < 0.02$ vs $m$), 
presumably due to fatigue. 
Adding concept name (penguin/journalist) did not improve model fit, nor did
addition of any interaction term. Thus participants produced equal numbers of
features for penguin and journalist, benefited equally from hinting for
both concepts, and showed equal decrements in performance when appearing in
the second block. 

\begin{table}
  \centering
  \begin{tabular}{lrrrrrrrr}
      \hline
      & npar & AIC & BIC & logLik & deviance & $\chi^2$ & df & Pr\\
      \hline
      $m_0$ & 3 & 602.5 & 609.4 & -298.2 & 596.5 & - & - & -\\
      $m$ & 4 & 592.4 & 601.5 & -292.2 & 584.4 & 12.1 & 1 & 0.00047   \\
      \hline
      $m$ & 4 & 592.4 & 601.5 & -292.2 & 584.4 & - & - & - \\
      $m_1$ & 5 & 594.2 & 605.6 & -292.1 & 584.2 & 0.1731 & 1 & 0.677 \\
      $m_2$ & 6 & 595.7 & 609.3 & -291.8 & 583.7 & 0.6994 & 2 & 0.704 \\
      $m_3$ & 5 & 588.6 & 600.0 & -289.3 & 578.6 & 5.756 & 1 & 0.016 \\
      \hline
      $m_3$ & 5 & 588.6 & 600.0 & -289.3 & 578.6 & - & - & - \\ 
      $m_4$ & 6 & 590.5 & 604.1 & -289.2 & 578.5 & 0.1301 & 1 & 0.718 \\ 
      $m_5$ & 10 & 596.3 & 619.1 & -288.1 & 576.3 & 2.277 & 5	& 0.809 \\
      \hline
    \end{tabular}
    \caption{Model comparison of null model $m_0$, model with condition $m$,
      model with condition and concept $m_1$, model with condition-concept
      interaction $m_2$, model with condition and block $m_3$. Model $m_4$ adds
      an interaction between condition and block, while $m_5$ is the full-model
      with all interactions.}
    \label{tab:model_comp}
\end{table}

\textbf{Do participants produce more diverse information?} 
Perhaps the hinting effect just described does not reflect improved
creativity, but just leads participants to generate additional phrases
composed of words and ideas already used. For instance, after generating
{\it has feathers} and {\it is black} for penguin, then receiving a hint,
participants may feel obligated to write new information even if no
new ideas have arisen, and so may generate a new phrase that restates
prior features (e.g. {\em has black
feathers}).

To evaluate this possibility, we evaluated the the number of {\it word types} produced. 
Taking all unique words produced, we performed case-folding, stopword removal, stemming, and
lemmatization on the participant's responses to compute the total of word types
produced in each condition. Participants produced a median of 57 word types in the hinted
condition and 42 in the unhinted condition. A mixed effects model regressing
word type count per participant on the experimental condition shows that this
result is significant at $p < 0.01$. 

We further considered word type density in each condition---the number of
word types divided by the total word count in a participant's responses. The median density
in the hinted condition was 0.56 while in the unhinted condition 0.59, 
a difference that is not statistically reliable. Thus the increase in productivity with hinting
was not accompanied by a decrease in word density--that is.

\textbf{Do hints inspired semantically related ideas?} To assess whether
features produced int he hinted condition are inspired by the most recent hint,
we computed a minimum-linkage distance between hints and features.

\begin{figure}[htbp]
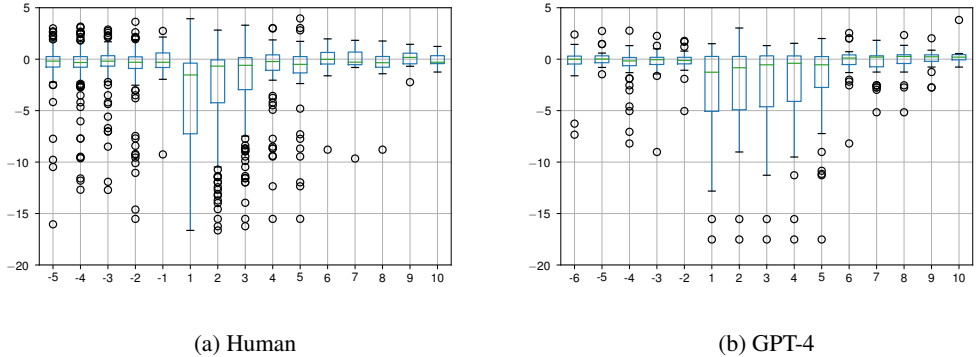

  \centering
  \begin{subfigure}[b]{0.49\textwidth}
    \centering
    \resizebox{1\textwidth}{!}{\input{figs/q3.b.pgf}}
    \caption{Human}
    \label{fig:figure1}
  \end{subfigure}
  \begin{subfigure}[b]{0.49\textwidth}
    \centering
    \resizebox{1\textwidth}{!}{\input{figs/q3.llm.pgf}}
    \caption{GPT-4}
    \label{fig:figure2}
  \end{subfigure}
  \caption{Z-score of distance between $i$-th feature and mean distance to hint before and after hint.}
  \label{fig:both_figures}
\end{figure}

That is, considering distance
$$d(x,y) = \min_{v \in x} \min_{w \in y} \|\phi(v),\phi(w)\|_2$$
for sentence $x$ and hint $y$, where $\phi(\cdot)$ takes word embeddings, we can
consider the empirical distribution of distances between hint $y$ and
concept $c$ over all sentences produced by all subjects in the unhinted
condition for concept $c$. This distance is negative when when a feature
produced is more semantically related to a hint than is the mean feature. 
\textbf{Figure~\ref{fig:figure1}} shows this metric normalized as z-scores
for features produced prior and subsequent to the request for a hint. 
Words leading up to the hint are not more semantically related to the
upcoming hint words than average, but following the hint, features produced become
much more related to the hint words--an effect that wanes after about
five produced features.

\textbf{Does the bandit learn to prefer some strategies over others?} As described in section
3.2, there is a rationale under which each hinting strategy could potentially help
human participants ``break out'' of their mental block to produce more information
in the feature listing task. A key question, then, is whether the bandit algorithm
reliably learns to prefer some strategies over others. Such a preference might
indicate, for instance, whether human behavior in the task is best enhanced by
expanding semantic neighborhoods already offered (semantic strategy), ``jumping'' to
a randomly-selected low-frequency word (frequency strategy), or broadly and systematically covering 
unexplored parts of semantic space (diversity strategy).

To answer this question, we first considered which arm had the lowest loss at the
end of a participant's session. For 17 participants, there was a unique least-loss arm; among
these, the semantic strategy won in cases, the frequency strategy in 5, and the diversity strategy in 2. The semantic strategy won more often than expected by chance if all arms have equal probability of winning ($p < 0.03$ given 1 in 3 chance of winning). 

We next computed the mean weight on each arm at the end of the session for each participant. The semantic arm had a mean weight of 0.40, reliably higher than both the frequency arm (mean = 0.31, $p < 0.05$ paired-samples t-test) and the diversity arm (mean = 0.29, $p < 0.03$ paired-samples t-test). Thus on average across participants the bandit prefers the semantic arm.

\begin{figure}
  \centering
  \includegraphics[width=\textwidth]{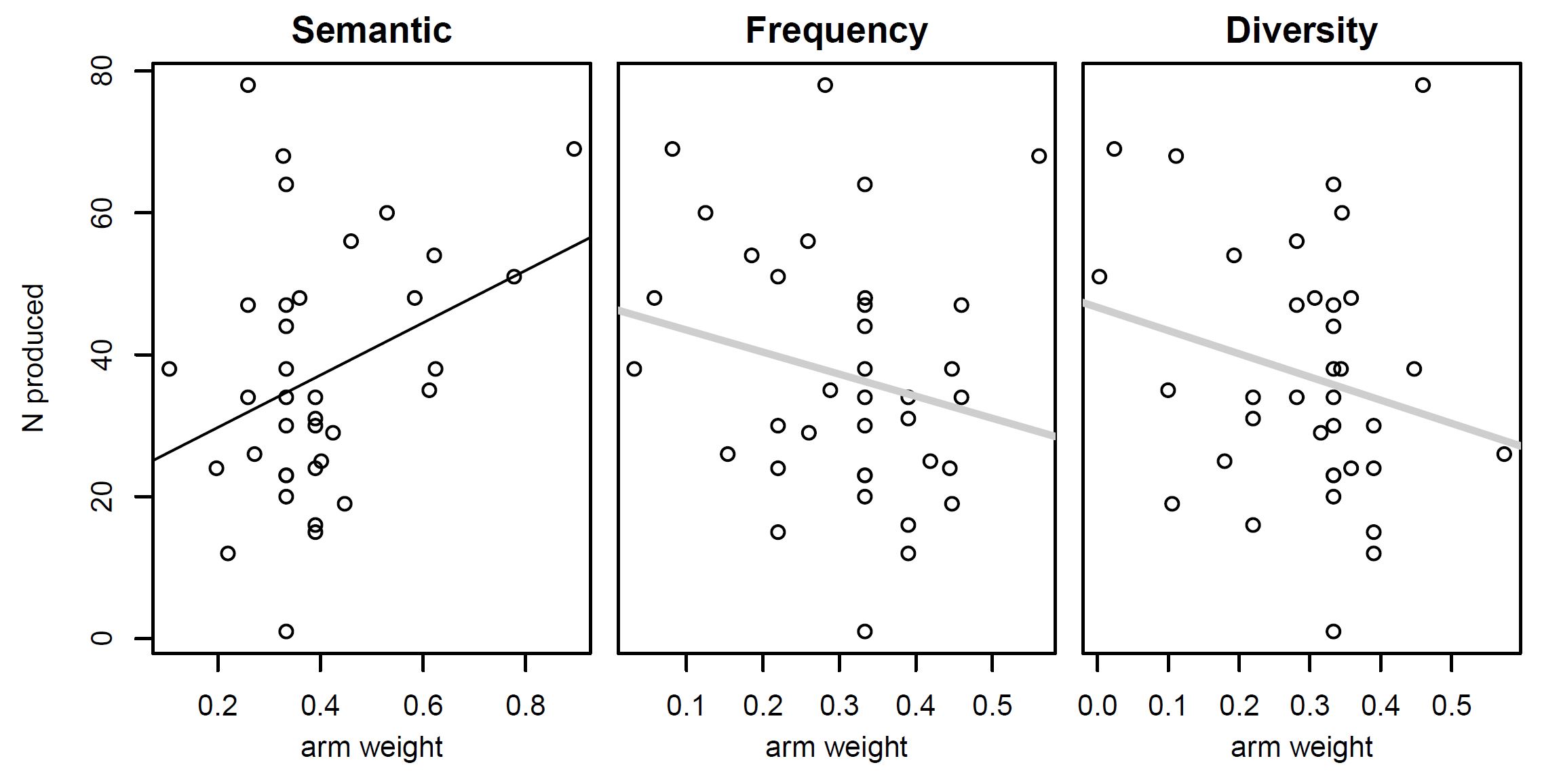}
  \caption{Relation between final weight on each arm and number of responses produced for all participants. For the semantic arm, weight magnitude correlates with better performance; the relationship is not reliable for the other
  two arms.}
  \label{fig:weight_performance}
\end{figure}

Finally, we considered how the final weights learned on each arm related to the number of participants produced by each participants. Results are shown in Figure \ref{fig:weight_performance}. When bandits learned larger weights on the semantic arm, the corresponding participant produced more features ($r = 0.33$, $p < 0.05$). No such relationship was observed for either of the other arms.

Together this evidence suggests that the improvement in performance shown above is best driven by the semantic hinting strategy, and that the MAB is capable of discovering the best strategy from human behavior.

\section{Study 2: Multi-Armed Bandit Interacting with GPT-4}

Study 1 showed that MABs can find effective strategies for improving human performance in a cognitive task, at least when selecting from amongst a small set of possibilities. Yet for many problems the space of possible hinting strategies may be very large, making it infeasible to train the bandit directly on human behavior. Study 2 assessed whether GPT-4, a large language model, exhibits similar patterns of behavior, and hence whether it can be used as a proxy for training the MAB. At first blush this possibility might seem unlikely: LLMs like GPT-4 are not transparently subject to the same cognitive constraints as humans, and it is not obvious that they will get ``stuck'' in simple tasks like feature listing. Yet several recent studies have suggested that such models often exhibit behaviors remarkably similar to those of human participants in a host of different language-based tasks (see \cite{dillion2023can} for review). We therefore conducted a model analog of Study 1 but using GPT-4 to simulate human behavior. As with the human study, the model was instructed to ask for hints when it ran out of ideas, and its responses were fed to the same MAB algorithm, which then generated hints according to the same procedures described earlier. LLM responses were tabulated and analyzed in the same way as human data. 

 \subsection{Methods.}

\begin{table}[htbp]
   \caption{Prompts used on GPT-4}
   \label{tab:prompt_variations}
   \begin{tabularx}{\textwidth}{|c|X|X|}
     \hline
     \textbf{\_} & \textbf{Unhinted} & \textbf{Hinted} \\ \hline
     \textbf{Initial Prompt} & Please type as many properties of journalist as you can think of. If you think you have exhausted all ideas, say "Give Up". Please use the format below. 1. [PROPERTY 1] \newline 2. [PROPERTY 2] & Please type as many properties of journalist as you can think of. When you run out of ideas, ask for a hint by saying 'Get Hints'. If you think you have exhausted all ideas, say 'Give Up'. Please use the format below. 1. [PROPERTY 1] \newline 2. [PROPERTY 2] \\ \hline
     \textbf{Subsequent Prompt} & N/A & Here are some hints: \emph{Hint1, Hint2, Hint3, Hint4, Hint5}. If they are not helpful, ask for another hint by saying 'Get Hints'. If you have exhausted your knowledge, say 'Give Up'. \\ \hline
\end{tabularx}
\end{table}

\begin{figure}[ht]
 \centering
 \includegraphics[width=\textwidth]{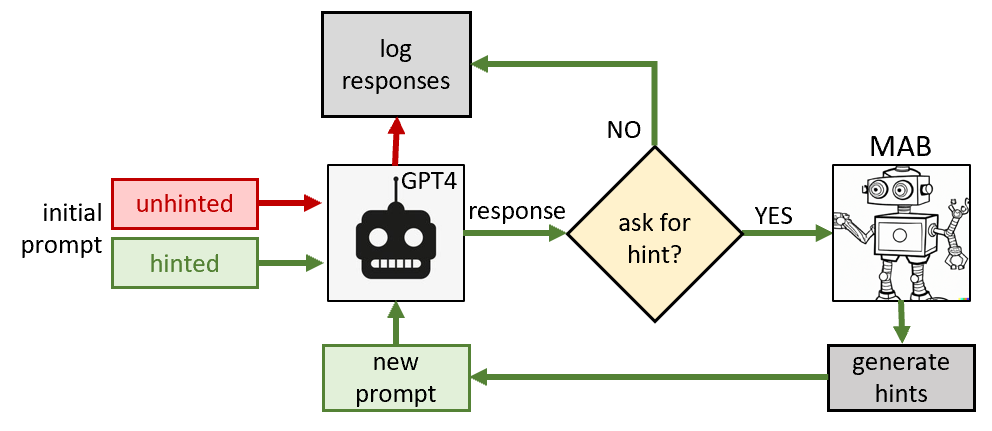}
 \caption{Control flow for GPT-4 study. Red arrows indicate unhinted condition, green arrows the hinted condition.}
 \label{fig:workflow}
\end{figure}

The GPT-4 experiments were run in early May 2023 using the OpenAI API, with the default temperature of 1 to ensure variability of responses akin to variation across individual participants. In lieu of the web interface from human experiments, we used prompts to guide the model's behavior as illustrated in Table \ref{tab:prompt_variations}. An experiment started by instructing the LLM with the initial prompt for the corresponding condition (hinted/unhinted) and concept (penguin/journalist). The control flow for simulating the 'Unhinted' and 'Hinted' conditions is shown in Figure \ref{fig:workflow}. In both conditions, the model was instructed to give up when it had exhausted its knowledge. Critically, in the hinted condition the AI was additionally instructed that it could ask for hints when it ran out of ideas. Of course, these instructions don't ensure that the model has indeed run out of ideas, or is in need of a hint--rather, the prompts provide a linguistic context that may lead the LLM to behave similarly to a human participant that indeed may run out of ideas or need hints. The GPT-4 experiment was run 60 times, simulating 15 participants in each cell of the condition (hinted/unhinted) by concept (penguin/journalist) design. Data were analyzed identically to the human studies with the sole exception that we used standard between-subjects regression models rather than mixed-effects models (since there is no model analog to a within-subjects manipulation).

\subsection{Results.}

\textbf{Does GPT-4 show human-like behavior with and without hints?} 
GPT-4's output is intrinsically limited only by the pre-specified size of its
output buffer--the model lacks the cognitive limitations that lead humans to get ``stuck.''
Nevertheless, GPT-4 limited its output in both hinted and unhinted conditions,
ceasing to produce new features well before reaching the specified output
token buffer size. That is, the instruction to give up when its knowlege
was exhausted led the model to produce comparable numbers of features to humans.
Across the two concepts, a median of 31 features were produced in the hinted condition (compare to 34 for human)
versus 25 in the unhinted condition (compare to 27 for human).

To assess whether the model produced more items in the hinted condition, note
that the analysis is simpler here as there is no meaningful mixed effect and no
concern about block effect. Thus ANOVA results are shown in
\textbf{Table~\ref{tab:model_comp_gpt}}. As in the human participant experiment,
the model produced reliably more features in the hinted condition $p < 0.001$. 
Adding concept name (penguin/journalist) did not improve model fit, with or without
the interaction term, indicating that, like human participants, the model produced similar
numbers of features for both concepts. 

Since the model receives additional prompt words when it asks for a hint, it may not seem surprising
that it subsequently produces more information. Note, however, there is no requirement for
the model to ask for a hint at all, or even to limit the number of its responses.
Nevertheless, the model does regularly ask for hints when given the opportunity to do so,
allowing it to then mimic the human behavior of requesting and benefiting hints provided.

\begin{table}
  \centering
  \begin{tabular}{lrrrrrr}
    \hline
    & RSS & df & Sum of Sq & F & Pr($>$F) \\
    \hline
    $m_0$ & 2838.2 & & & & \\
    $m$  & 2292.2 & 1 & 546.02 & 13.816 & 0.0004558 \\
    \hline
    $m$ & 2292.2 & & & & \\
    $m_1$ & 2177.3 & 1 & 114.82 & 3.0057 & 0.08838 \\
    \hline
    $m$ & 2292.2 & & & & \\
    $m_2$ & 2174.5 & 2 & 117.63 & 1.5147 & 0.2287 \\
    \hline
  \end{tabular}
  \caption{Model comparison of null model $m_0$, model with condition $m$,
    model with condition and concept $m_1$, model with condition-concept
    interaction $m_2$.}
  \label{tab:model_comp_gpt}
\end{table}

\textbf{Does GPT-4 produce more word types in hinted conditions?}

GPT-4 produced a median of 95 word tokens in the hinted condition and 83 in the
unhinted conditions, with ANOVA showing this to be a statistically reliable difference 
with $p < 0.05$. Thus, like humans, GPT-4 does not simply ``remix'' words
already produced when provided with hints.

The median word type density in the hinted condition was 0.60 and 0.63 in the
unhinted condition. The small difference is comparable to that observed in human
participants, but in contrast to Study 1, ANOVA results showed this to be a statistically reliable effect
$p < 0.05$. Thus GPT-4s responses become reliably less diverse with hints.

\textbf{Is GPT-4 inspired by the hints it receives?} As in study 1, we considered
whether the words generated by GPT-4 following a hint were semantically related
to the hints provided, relative to the mean similarity to produced words in the
unhinted condition. \textbf{Figure~\ref{fig:figure2}} shows the results, plotted 
in the same manner as Study 1. We observed a similar pattern to human behavior,
in which responses provided immediately following a hint were more semantically
related to the hint words than those immediately preceding the hint--an effect that 
again persisted over about 5 subsequent features.

\textbf{Do bandits learned the same arm preferences from GPT behavior?} 
The preceding analyses suggest that GPT-4's behavior in the task is remarkably
similar to human behavior in many respects. Does this mean that a bandit
trained to ``boost'' GPT-4's performance will select similar strategies 
to those trained on human data?

Although GPT-4 asked for fewer hints than humans overall (median of 4 vs 8),
nevertheless the same best strategy (semantic) was identified as the
preferred arm, accounting for 65\% of arm pulls in the GPT-4
sessions. As with the human-trained bandits,the final arm probability 
for the semantic strategy was reliably higher than both the frequency-based
strategy ($p < 0.002$ paired-sample t-test) and the diverse exploration
strategy ($p < 0.01$). In other words, had we chosen a strategy based
on GPT-4 simulated human behavior, we would have selected the arm shown
in Study 1 to best improve human performance.

\section{Conclusion and Future Work}

Given recent breakthroughs in generative AI, it is not surprising that there is
curiosity about the potential for human-AI interaction to enhance creativity.
Our results suggests that bandit AI agents can reliably improve human
performance on a feature generation task known to reliably produce mental
roadblocks. Though we focus on three simple strategies, a benefit of the
proposed bandit framework is its scalability to a much larger set of possible
prompting strategies. It is easy to see how the current approach could be deployed
on crowd-sourced human judgments to adjudicate a wide range of hypotheses about
ways of improving human creativity across a variety of more challenging tasks.

A second contribution of our work is the validation of
LLMs as a proxy for a human participant for this task. The remarkably human-like
behavior of GPT-4 is somewhat surprising given that the models are not subject 
to processing constraints that limit human behavior. Indeed, GPT-4 can potentially 
generate a very extensive list of properties for a given concept in very little
time. It is remarkable that, when given the option of asking for help or giving up,
the model shows human-like tendencies in the information it produces.
We further showed that this isomorphism with human behavior allows the bandit
algorithm to adjudicate prompting strategies likely to work with people solely by
interacting with the language model.

Taken together, these two
contributions suggest a tantalizing direction for future work: using LLMs to
efficiently evaluate a much broader set of prompting strategies. A key goal for
future research will be to extend the range of potential creativity applications
beyond the feature recall task to reshape our understanding of human-AI
co-creativity.

This project is supported in part by NSF grants 1545481,
1704117, 1836978, 2023239, 2041428, 2202457, ARO
MURI W911NF2110317, AF CoE FA9550-18-1-0166, and Intuit.

\bibliographystyle{plain}
\bibliography{arxiv}

\clearpage

\appendix

\section{Supplementary Materials}

\begin{figure}[ht]
  \centering
  \frame{\includegraphics[width=1\textwidth]{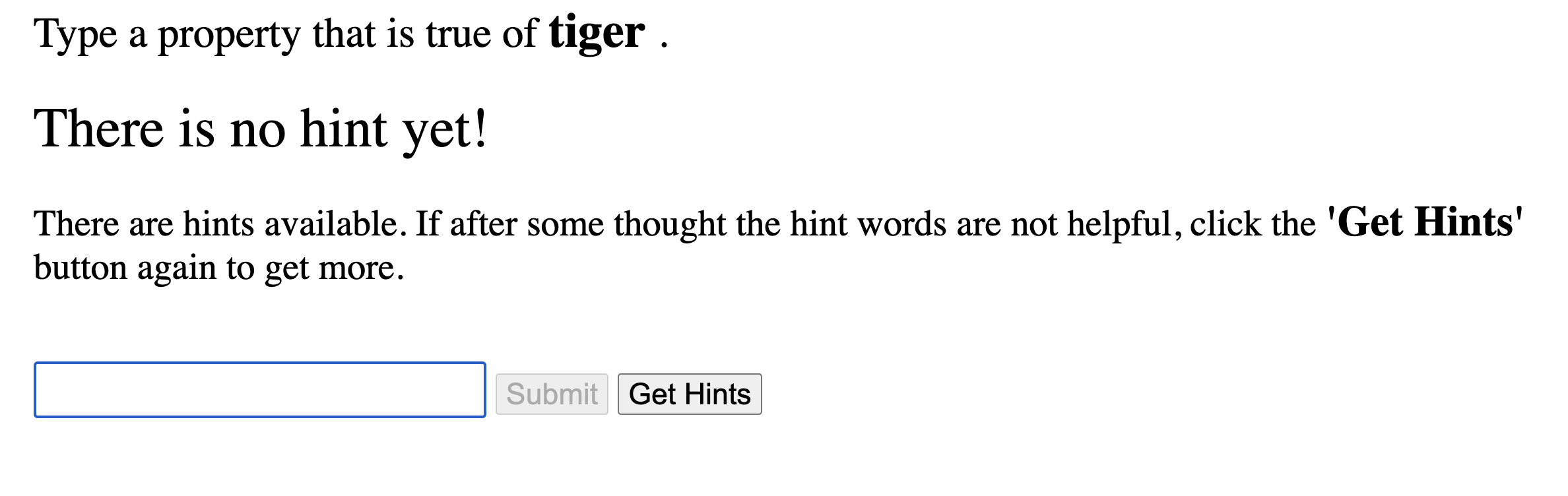}}
  \caption{Screenshot of the user interface.}
  \label{fig:interface}
\end{figure}

\begin{figure}[ht]
  \centering
  \frame{\includegraphics[width=1\textwidth]{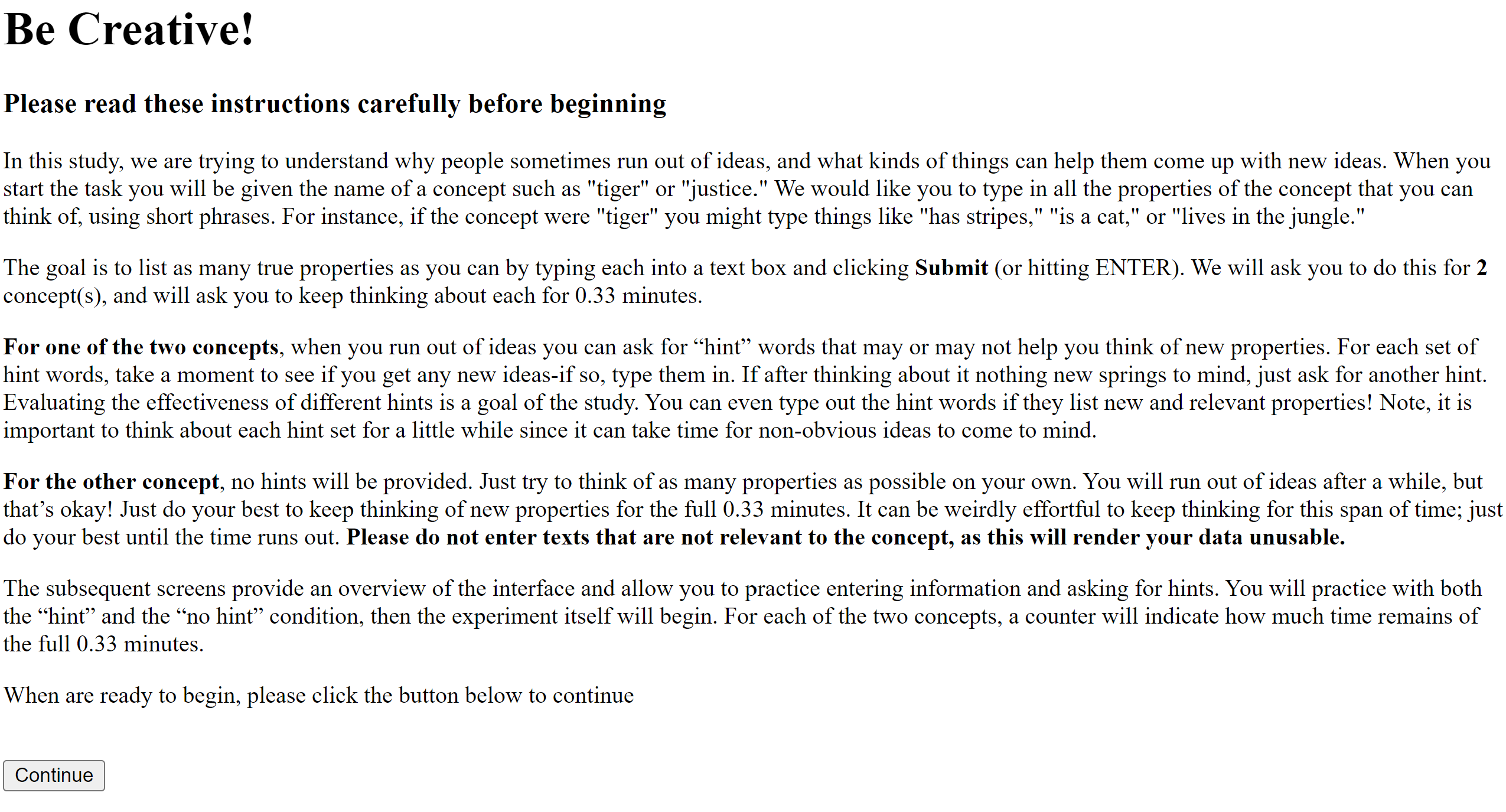}}
  \caption{Screenshot of the instruction page.}
  \label{fig:instructions}
\end{figure}

\subsection{Arm Specification}
The general input required by different arms is as follows: 
\begin{itemize}
    \item $VOCAB$: the set of all the words.
    \item $Candidates$: the set of candidate words (used for selecting prompts).
    \item $Said$: the set of user-entered words intersect $VOCAB$.
    \item $Heard$: the set of words from prompts. If the concept word tuple is in $VOCAB$, $Heard$ would be initialized to contain it as the user is exposed to the concept word once the task starts. 
    \item $d$: the Euclidean distance function that measures the distance of two-word embeddings from tuples in $VOCAB$.
    \item $b$: the number of prompts generated each time.
\end{itemize}

In order to avoid duplication, we removed the logic when the size of $Candidates$ is less than $b$ in the following pseudocode. In that case, each arm will simply return $Candidates$ directly.

\subsubsection{Arm One}

\begin{algorithm}[H]
 \hspace*{\algorithmicindent} \textbf{Input} Candidates, b\\
 \hspace*{\algorithmicindent} \textbf{Output} a list of word types with size at most $b$
   \caption{Arm One - Frequency hints with candidates}
    \begin{algorithmic}[1]
      \Function{frequencyHints}{$Candidates, b$}

            \If{size of $Candidates$ is less than $b$}
            
                \State \Return $Candidates$
                
            \EndIf
            
            \State $hints$ $\sim$ Multinomial($Candidates, b$) \Comment{the probability of sampling each word is the frequency in the sample space $Candidates$}
            
            \State \Return $hints$
            
       \EndFunction
\end{algorithmic}
\end{algorithm}

\subsubsection{Arm Two}
\begin{algorithm}[H]
 \hspace*{\algorithmicindent} \textbf{Input} Candidates, Said, Heard, d, b, threshold\\
 \hspace*{\algorithmicindent} \textbf{Output} a list of word types with size at most $b$
   \caption{Arm Two - efficientDiversitySampling}
    \begin{algorithmic}[1]
      \Function{frequencyHints}{$Candidates, Said, Heard, d, b, threshold$}

            \State $hints = \varnothing$
            
            \If{size of $Candidates$ is greater than $threshold$}
            
                \State $C=$ uniformly sample $threshold$ word types from $Candidates$
            
            \Else

                \State $C=Candidates$
                
            \EndIf
                       
            \State $Known=Said \cup Heard$

            \If{$Known = \varnothing$ and $C \not=\varnothing$}

                \State $Known = \{ \text{a word uniformly chosen from }C\}$

            \EndIf
            
            \While{$C \not = \varnothing$ and $b>0$ }
            
                \State $m = \text{size of }C$
                
                \State \Comment{for each word $w$ with index $i$ in $C$, assign $d_i$ as the sample density, where $d_i = (getWordSetDistance(w, Known, d))^2$}
                
                \State $p_i = \dfrac{d_i}{\sum_{j=1}^{m}d_j} $ for $i=1, \dots, m$
                
                \State $i \sim $ MultiNomial($[p_1, \dots, p_m]$, 1) \Comment{$w$ is the word type with index $i$}

                \State $C = C - \{w\}$
                
                \State $Known = Known \cup \{w\}$

                \State $hints = hints \cup \{w\}$

                \State $b = b - 1$
            \EndWhile
                        
            \State \Return $hints$
            
       \EndFunction
\end{algorithmic}
\end{algorithm}

Note: 
\begin{itemize}
    \item $getWordSetDistance$ is defined as the word set distance in the explanation section of this arm. 
    \item Although we use set notation for $hints$, the display order to the user is the same as the order in which the hints are generated.
\end{itemize}
\newpage

\subsubsection{Arm Three}
\begin{algorithm}[H]
 \hspace*{\algorithmicindent} \textbf{Input} Candidates, Said, d, removedFromSaid, b\\
 \hspace*{\algorithmicindent} \textbf{Output} a list of word types with size at most $b$
   \caption{Arm Three - Least-frequency hints near Said}
    \begin{algorithmic}[1]
      \Function{leastFrequencyHintsNearSaid}{$Candidates, Said, d, removedFromSaid, b$}

            \State $priorityQueue = Said \backslash removedFromSaid$ \Comment{Low frequency preferred priority queue}

            \If{$priorityQueue = \varnothing$ }
                \State return $\varnothing$
            \EndIf

            \State $s = \text{pop}(priorityQueue)$ 
            
            \State $removedFromSaid = removedFromSaid \cup \{s\}$

            \State $hints = \text{kNN}(s, Candidates, b, d)$
       \EndFunction
\end{algorithmic}
\end{algorithm}

Note:
\begin{itemize}
    \item $removedFromSaid$ is a persistent variable across different calls to the above arm three. But it is only persistent for only one participant. 
    \item kNN is the k-nearest-neighbor algorithm, and parameter b plays the role of k.
\end{itemize}

\end{document}